\documentclass[11pt,a4paper]{article}
\usepackage[hyperref]{AACL-IJCNLP2020}
\usepackage{times}
\usepackage{latexsym}
\usepackage{graphicx}
\usepackage{booktabs}
\usepackage{CJKutf8}
\usepackage{bbding}
\usepackage{pifont}
\usepackage{amssymb}
\usepackage{wasysym}
\usepackage{caption}
\usepackage{subcaption}

\usepackage{microtype}

\aclfinalcopy 

\makeatletter
\newcommand{\thickhline}{%
    \noalign {\ifnum 0=`}\fi \hrule height 1pt
    \futurelet \reserved@a \@xhline
}

\title{Mixed-Lingual Pre-training for Cross-lingual Summarization}

\author{Ruochen Xu\thanks{$\;\;$Equal contribution}$\;$, Chenguang Zhu$^*$, Yu Shi, Michael Zeng, Xuedong Huang \\
Microsoft Cognitive Services Research Group\\
  \texttt{\{ruox,chezhu,yushi,nzeng,xdh\}@microsoft.com} \\}

\date{}

\begin{document}
\maketitle
\begin{abstract}
Cross-lingual Summarization (CLS) aims at producing a summary in the target language for an article in the source language. Traditional solutions employ a two-step approach, i.e. translate$\rightarrow$summarize or summarize$\rightarrow$translate. Recently, end-to-end models have achieved better results, but these approaches are mostly limited by their dependence on large-scale labeled data. We propose a solution based on mixed-lingual pre-training that leverages both cross-lingual tasks such as translation and monolingual tasks like masked language models. Thus, our model can leverage the massive monolingual data to enhance its modeling of language. Moreover, the architecture has no task-specific components, which saves memory and increases optimization efficiency. We show in experiments that this pre-training scheme can effectively boost the performance of cross-lingual summarization. In Neural Cross-Lingual Summarization (NCLS) \citep{ncls} dataset, our model achieves an improvement of 2.82 (English to Chinese) and 1.15 (Chinese to English) ROUGE-1 scores over state-of-the-art results.
\end{abstract}

\section{Introduction}
Text summarization can facilitate the propagation of information by providing an abridged version for long articles and documents. Meanwhile, the globalization progress has prompted a high demand of information dissemination across language barriers. Thus, the cross-lingual summarization (CLS) task emerges to provide accurate gist of articles in a foreign language.

Traditionally, most CLS methods follow the two-step pipeline approach: either translate the article into the target language and then summarize it \citep{mtsum}, or summarize the article in the source language and then translate it \citep{summt}. Although this method can leverage off-the-shelf summarization and MT models, it suffers from error accumulation from two independent subtasks. Therefore, several end-to-end approaches have been proposed recently \citep{ncls,naacl2019,acl2019}, which conduct both translation and summarization simultaneously. Easy to optimize as these methods are, they typically require a large amount of cross-lingual summarization data, which may not be available especially for low-resource languages. For instance, NCLS \citep{ncls} proposes to co-train on monolingual summarization (MS) and machine translation (MT) tasks, both of which require tremendous labeling efforts.

On the other hand, the pre-training strategy has proved to be very effective for language understanding \citep{bert,gpt} and cross-lingual learning \citep{xlm, aaai2020}. One of the advantages of pre-training is that many associated tasks are self-learning by nature, which means no labeled data is required. This greatly increases the amount of training data exposed to the model, thereby enhancing its performance on downstream tasks.

Therefore, we leverage large-scale pre-training to improve the quality of cross-lingual summarization. Built upon a transformer-based encoder-decoder architecture \citep{transformer}, our model is pre-trained on both monolingual tasks including masked language model (MLM), denoising autoencoder (DAE) and monolingual summarization (MS), and cross-lingual tasks such as cross-lingual masked language model (CMLM) and machine translation (MT). This mixed-lingual pre-training scheme can take advantage of massive unlabeled monolingual data to improve the model's language modeling capability, and leverage cross-lingual tasks to improve the model's cross-lingual representation. We then finetune the model on the downstream cross-lingual summarization task.

Furthermore, based on a shared multi-lingual vocabulary, our model has a shared encoder-decoder architecture for all pre-training and finetuning tasks, whereas NCLS \citep{ncls} sets aside task-specific decoders for machine translation, monolingual summarization, and cross-lingual summarization. 

In the experiments, our model outperforms various baseline systems on the benchmark dataset NCLS \citep{ncls}. For example, our model achieves 3.27 higher ROUGE-1 score in Chinese to English summarization than the state-of-the-art result and 1.28 higher ROUGE-1 score in English to Chinese summarization. We further conduct an ablation study to show that each pretraining task contributes to the performance, especially our proposed unsupervised pretraining tasks.
 
\section{Related Work}
\subsection{Pre-training} 
Pre-training language models \citep{bert, unilm} have been widely used in NLP applications such as question answering \citep{sdnet}, sentiment analysis \citep{elmo}, and summarization \citep{leadbias,yang2020ted}.
In multi-lingual scenarios, recent works take input from multiple languages and shows great improvements on cross-lingual classification \citep{xlm, pires2019multilingual, huang2019unicoder} and unsupervised machine translation \citep{liu2020multilingual}.
\citet{artetxe2019massively} employs the sequence encoder from a machine translation model to produce
cross-lingual sentence embeddings. \citet{aaai2020} uses multi-lingual pre-training to improve cross-lingual question generation and zero-shot cross-lingual summarization. Their model trained on articles and summaries in one language is directly used to produce summaries for articles in another language, which is different from our task of producing summaries of one language for an article from a foreign language.

\subsection{Cross-lingual Summarization}
Early literatures on cross-lingual summarization focus on the two-step approach involving machine translation and summarization \citep{mtsum,summt}, which often suffer from error propagation issues due to the imperfect modular systems. Recent end-to-end deep learning models have greatly enhanced the performance. \citet{IEEE2018} presents a solution to zero-shot cross-lingual headline generation by using machine translation and summarization datasets. \citet{acl2019} leverages monolingual abstractive summarization to achieve zero-shot cross-lingual abstractive sentence summarization. NCLS \citep{ncls} proposes a cross-lingual summarization system for large-scale datasets for the first time. It uses multi-task supervised learning and shares the encoder for monolingual summarization, cross-lingual summarization, and machine translation. However, each of these tasks requires a separate decoder. In comparison, our model shares the entire encoder-decoder architecture among all pre-training and finetuning tasks, and leverages unlabeled data for monolingual masked language model training. A concurrent work by \citet{zhu2020attend} improves the performance by combining the neural model with an external probabilistic bilingual lexicon.
\section{Method}
\begin{CJK*}{UTF8}{gbsn}
\begin{table*}[ht]
\centering
\resizebox{\textwidth}{!}{%
\begin{tabular}{lcccc}
\toprule
\textbf{Objective} & \textbf{Supervised} & \textbf{Multi-lingual} & \textbf{Inputs} & \textbf{Targets} \\ \midrule
Masked Language Model & & & France \textless{}X\textgreater{} Morocco in \textless{}Y\textgreater{} exhibition match. & \textless{}X\textgreater{} beats \textless{}Y\textgreater{} an \\ \midrule
Denoising Auto-Encoder & & &  France beats \textless{}M\textgreater{} in \textless{}M\textgreater{} exhibition . & France beats Morocco in an exhibition match. \\ \midrule
Monolingual Summarization & \checkmark &  & \begin{tabular}[c]{@{}c@{}}World champion France overcame a stuttering \\ start to beat Morocco 1-0 in a scrappy exhibition \\ match on Wednesday night.\end{tabular} & France beats Morocco in an exhibition match. \\ \midrule
Cross-lingual MLM & \checkmark & \checkmark  & \begin{tabular}[c]{@{}c@{}}France \textless{}X\textgreater{} Morocco in \textless{}Y\textgreater{} exhibition match. \\ 法国队在一场表演赛中击败摩洛哥队。\end{tabular} & \textless{}X\textgreater{} beats \textless{}Y\textgreater{} an \\ \midrule
Cross-lingual MLM & \checkmark & \checkmark & \begin{tabular}[c]{@{}c@{}}France beats Morocco in an exhibition match.   \\ \textless{}X\textgreater{}队在一场表演赛中\textless{}Y\textgreater{}摩洛哥队。\end{tabular} & \textless{}X\textgreater法国 \textless{}Y\textgreater击败 \\ \midrule
Machine Translation & \checkmark & \checkmark  & France beats Morocco in an exhibition match. & 法国队在一场表演赛中击败摩洛哥队。 \\ \bottomrule
\end{tabular}%
}
\caption{Examples of inputs and targets used by different objectives for the sentence ``France beats Morocco in an exhibition match'' with its Chinese translation. We use \textless{}X\textgreater{} and \textless{}Y\textgreater{} to denote sentinel tokens and \textless{}M\textgreater{} to denote shared mask tokens.}
\label{tab:obj-examples}
\end{table*}
\end{CJK*}

\subsection{Pre-training Objectives}
We propose a set of multi-task pre-training objectives on both monolingual and cross-lingual corpus. For monolingual corpus, we use the masked language model (MLM) from \citet{t5}. The input is the original sentence masked by sentinel tokens, and the target is the sequence consists of each sentinel token followed by the corresponding masked token. The other monolingual task is the denoising auto-encoder (DAE), where the corrupted input is constructed by randomly dropping, masking, and shuffling a sentence and the target is the original sentence. Since our final task is summarization, we also include monolingual summarization (MS) as a pre-training task.

To leverage cross-lingual parallel corpus, we introduce the cross-lingual masked language model (CMLM). CMLM is an extension of MLM on the parallel corpus. The input is the concatenation of a sentence in language A and its translation in language B. We then randomly select one sentence and mask some of its tokens by sentinels. The target is to predict the masked tokens in the same way as MLM. Different from MLM, the masked tokens in CMLM are predicted not only from the context within the same language but also from their translations in another language, which encourages the model to learn language-invariant representations. Note that CMLM is similar to the Translation Language Model (TLM) loss proposed in \citet{xlm}. The key differences are: 1) TLM randomly masks tokens in sentences from both languages, while CMLM only masks tokens from one language; 2) TLM is applied on encoder-only networks while we employ CMLM on the encoder-decoder network. In addition to CMLM, we also include standard machine translation (MT) objective, in which the input and output are the unchanged source and target sentences, respectively.

The examples of inputs and targets used by our pre-training objectives are shown in Table \ref{tab:obj-examples}. 

\subsection{Unified Model for Pre-training and Finetuning}
While NCLS \citep{ncls} uses different decoders for various pre-training objectives, we employ a unified Transformer \citep{transformer} encoder-decoder model for all pre-training and finetuning tasks. This makes our model learn a cross-lingual representation efficiently. A shared dictionary across all languages is used.  To accommodate multi-task and multilingual objectives, we introduce language id symbols to indicate the target language, and task symbols to indicate the target task. For instance, for the CMLM objective where the target language is Chinese, the decoder takes $<$cmlm$>$ and $<$zh$>$ as the first two input tokens.
We empirically find that our model does not suffer from the phenomenon of forgetting target language controllability as in \citet{aaai2020}, which requires manual freezing of encoder or decoder during finetuning. After pretraining, we conduct finetuning on cross-lingual summarization data.

\section{Experiments}
\begin{table*}[htbp]
\centering
\begin{tabular}{l|lll|lll}
\hline
 & \multicolumn{3}{l|}{English$\rightarrow$Chinese} & \multicolumn{3}{l}{Chinese$\rightarrow$English} \\ \hline
 & ROUGE-1 & ROUGE-2 & ROUGE-L & ROUGE-1 & ROUGE-2 & ROUGE-L \\ \hline
TETran & 26.15 & 10.60 & 23.24 & 23.09 & 7.33 & 18.74 \\
GETran & 28.19 & 11.40 & 25.77 & 24.34 & 9.14 & 20.13 \\
TLTran & 30.22 & 12.20 & 27.04 & 33.92 & 15.81 & 29.86 \\
GLTran & 32.17 & 13.85 & 29.43 & 35.45 & 16.86 & 31.28 \\
NCLS & 36.82 & 18.72 & 33.20 & 38.85 & 21.93 & 35.05 \\
NCLS-MS & 38.25 & 20.20 & 34.76 & 40.34 & 22.65 & 36.39 \\
NCLS-MT & 40.23 & 22.32 & 36.59 & 40.25 & 22.58 & 36.21 \\
XNLG & 39.85 & 24.47 & 28.28 & 38.34 & 19.65 & 33.66 \\ 
ATS & 40.68 & 24.12 & \textbf{36.97} & 40.47 & 22.21 & 36.89 \\ \hline
Ours & \textbf{43.50} & \textbf{25.41} & 29.66 & \textbf{41.62} & \textbf{23.35} & \textbf{37.26} \\ \hline
\end{tabular}%
\caption{ROUGE-1, ROUGE-2, ROUGE-L for English to Chinese and Chinese to English summarization on NCLS dataset.} \label{table:mainresult}
\end{table*}

\subsection{Dataset}
We conduct our experiment on NCLS dataset \citep{ncls}, which contains paired data of English articles with Chinese summaries, and Chinese articles with English summaries. The cross-lingual training data is automatically generated by a machine translation model.  For finetuning and testing, we followed the same train/valid/test split of the original dataset. We refer readers to Table 1 in \citet{ncls} for detailed statistics of the dataset.

For pre-training, we obtain monolingual data for English and Chinese from the corresponding Wikipedia dump. There are 83 million sentences for English monolingual corpus and 20 million sentences for Chinese corpus. For parallel data between English and Chinese, we use the parallel corpus from \citet{xlm}, which contains 9.6 million paired sentences.
For monolingual summarization objective, we use CNN/DailyMail dataset \citep{dailymail} for English summarization and LCSTS dataset \citep{hu2015lcsts} for Chinese summarization.

\subsection{Implementation Details}
Our transformer model has 6 layers and 8 heads in attention. The input and output dimensions $d_{model}$ for all transformer blocks are 512 and the inner dimension $d_{ff}$ is 2048. 

We use a dropout probability of 0.1 on all layers. We build a shared {S}entence{P}iece \citep{sentencepiece} vocabulary of size $33,000$ from a balanced mix of the monolingual Wikipedia corpus. The model has approximately 61M parameters.

For MLM we use a mask probability of 0.15. For DAE, we set both the mask and drop out rate to 0.1.
For all pre-training and finetuning we use RAdam optimizer \cite{radam} with $\beta_1=0.9$, $\beta_2=0.999$. The initial learning rate is set to $10^{-9}$ for pre-training and $10^{-4}$ for finetuning. The learning rate is linearly increased to $0.001$ with $16,000$ warmup steps followed by an exponential decay.
For decoding, we use a beam size of 6 and a maximum generation length of 200 tokens for all experiments.


\begin{table}[htbp]
\centering
\resizebox{0.47\textwidth}{!}{%
\begin{tabular}{@{}l|lll@{}}
\thickhline
 & \multicolumn{3}{c}{English$\rightarrow$Chinese} \\ \hline
 & ROUGE-1 & ROUGE-2 & ROUGE-L \\ \hline
Ours & 43.50 & 25.41 & 29.66 \\ \hline
-   MS & 42.48 & 24.45 & 28.49 \\
-   MT & 42.12 & 23.97 & 28.74 \\
-   MLM, DAE & 41.82 & 23.85 & 28.40 \\
-   All Pretraining & 41.12 & 23.67 & 28.53 \\ \thickhline
\end{tabular}%
}
\caption{Finetuning performance on English$\rightarrow$Chinese summarization starting with various ablated pre-trained models. 
}
\label{tab:ablation_study}
\end{table}

\subsection{Baselines}
We first include a set of pipeline methods from \citet{ncls} which combines monolingual summarization and machine translation.
\textbf{TETran} first translates the source document and then uses LexRank \citep{lexrank} to summarize the translated document. 
\textbf{TLTran} first summarizes the source document and then translates the summary.
\textbf{GETran} and \textbf{GLTran} replace the translation model in TETran and TLTran with Google Translator\footnote{https://translate.google.com/} respectively. 

We also include three strong baselines from \citet{ncls}: \textbf{NCLS}, \textbf{NCLS-MS} and \textbf{NCLS-MT}. NCLS trains a standard Transformer model on the cross-lingual summarization dataset. NCLS-MS and NCLS-MT both use one encoder and multiple decoders for multi-task scenarios. NCLS-MS combines the cross-lingual summarization task with monolingual summarization while NCLS-MT combines it with machine translation.

We finetune \textbf{XNLG} model from \citet{aaai2020} on the same cross-lingual summarization data. We finetune all layers of XNLG in the same way as our pretrained model.

Finally, we include the result of \textbf{ATS} from the concurrent work of \citet{zhu2020attend}.

\begin{figure*}
\centering
\begin{subfigure}{.5\textwidth}
  \centering
  \includegraphics[width=1.0\linewidth]{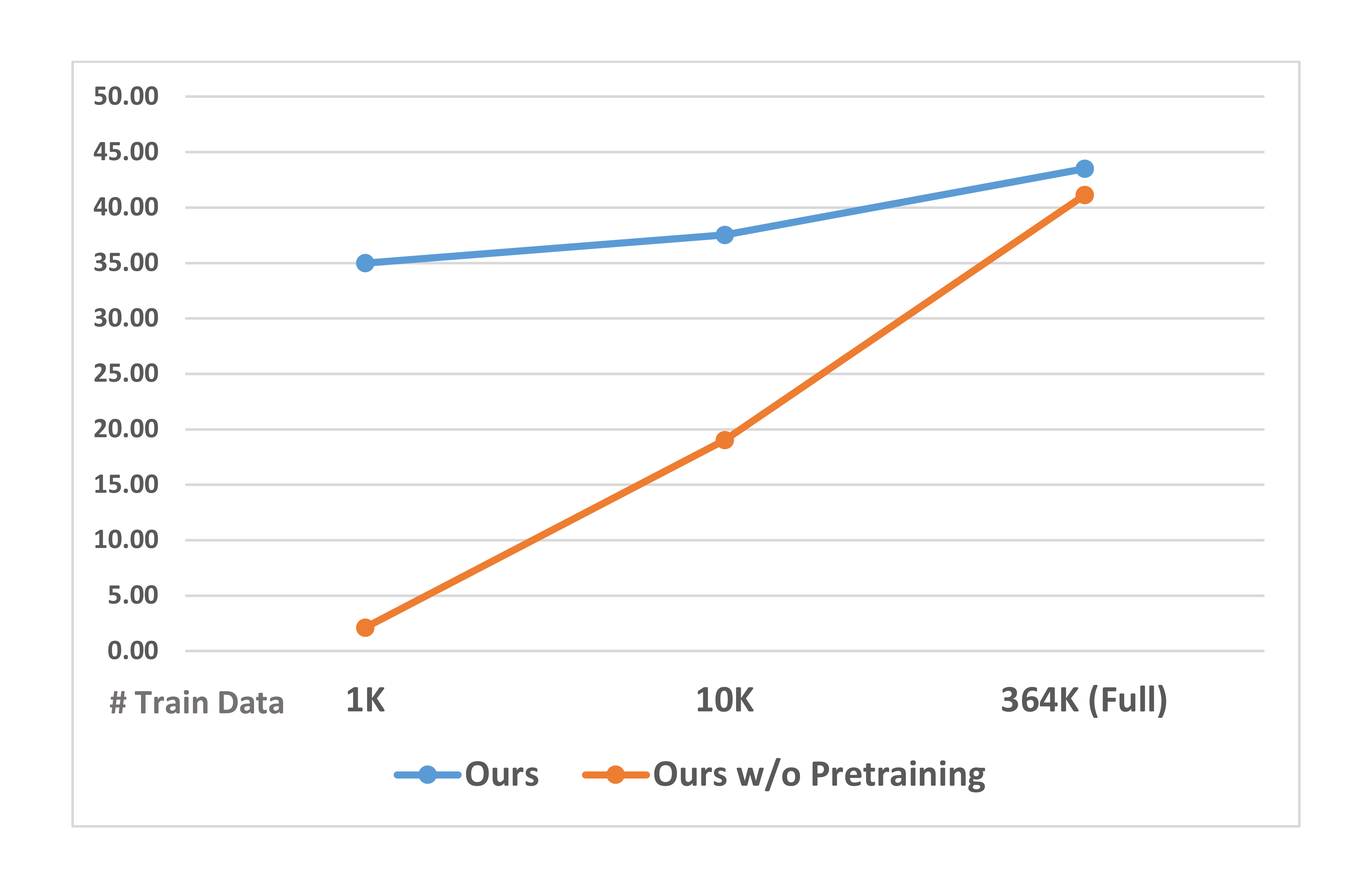}
  \caption{English$\rightarrow$Chinese ROUGE-1}
  \label{fig:low-resource-sub1}
\end{subfigure}%
\begin{subfigure}{.5\textwidth}
  \centering
  \includegraphics[width=1.0\linewidth]{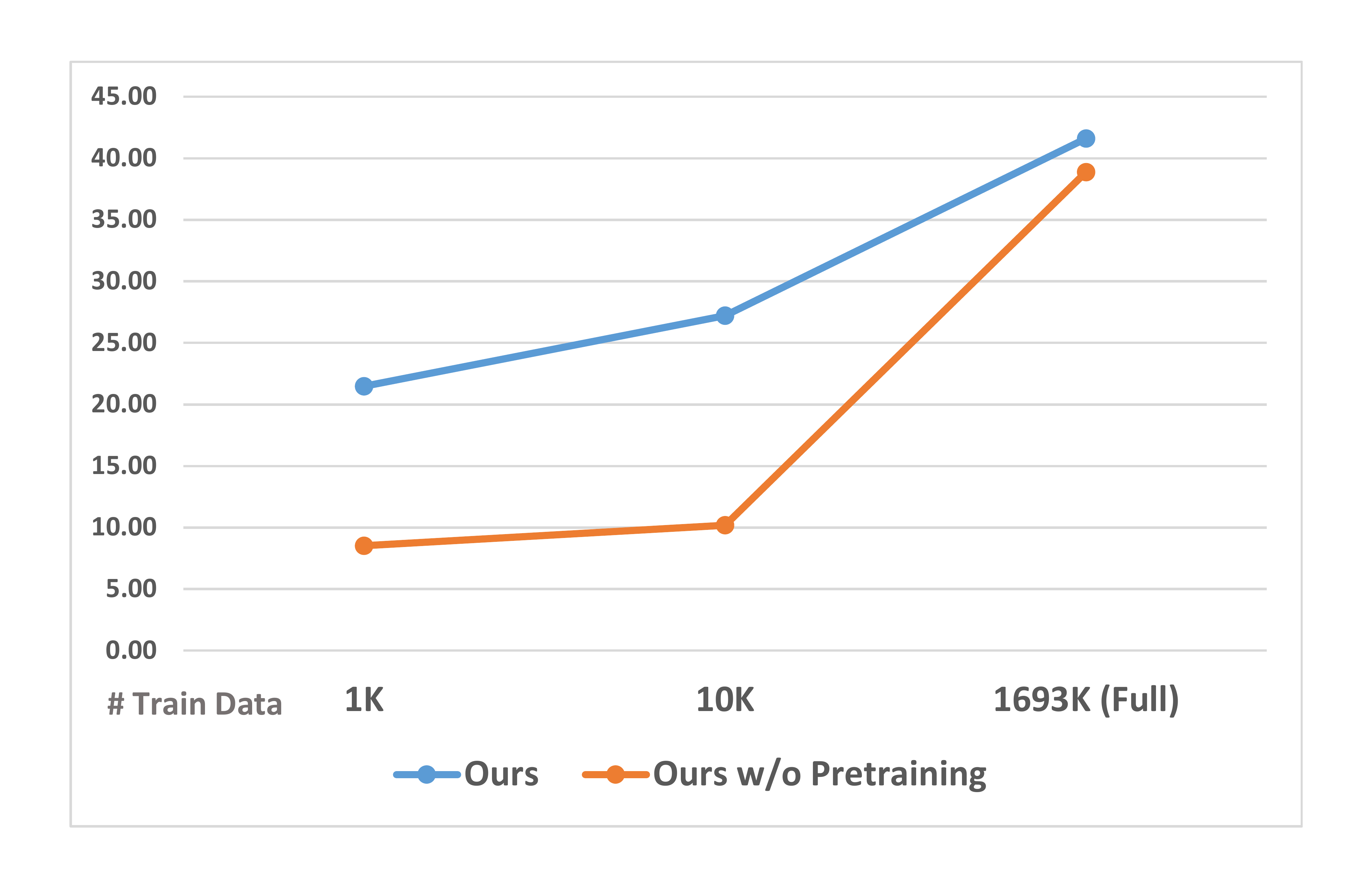}
  \caption{Chinese$\rightarrow$English ROUGE-1}
  \label{fig:low-resource-sub2}
\end{subfigure}
\caption{ROUGE-1 performance on NCLS dataset when the cross-lingual summarization training data is sub-sampled to size of 1k and 10k. The result on the full dataset is also shown.}
\label{fig:low-resource}
\end{figure*}

\subsection{Results}
Table \ref{table:mainresult} shows the ROUGE 
scores of generated summaries in English-to-Chinese and Chinese-to-English summarization. As shown, pipeline models, although incorporating state-of-the-art machine translation systems, achieve sub-optimal performance in both directions, proving the advantages of end-to-end models. 

Our model outperforms all baseline models in all metrics except for ROUGE-L in English-to-Chinese. For instance, our model achieves 2.82 higher ROUGE-1 score in Chinese to English summarization than the previously best result and 1.15 higher ROUGE-1 score in English to Chinese summarization, which shows the effectiveness of utilizing multilingual and multi-task data to improve cross-lingual summarization.

\subsection{Ablation Study}
Table \ref{tab:ablation_study} shows the ablation study of our model on English to Chinese summarization. We remove from the pre-training objectives i) all monolingual unsupervised tasks (MLM, DAE), ii) machine translation (MT), iii) monolingual summarization (MS), and iv) all the objectives. Note that "- All Pretraining" and NCLS both only train on the cross-lingual summarization data. The performance difference between the two is most likely due to the difference in model size, vocabulary, and other hyper-parameters.

As shown, the pre-training can improve ROUGE-1, ROUGE-2, and ROUGE-L by 2.38, 1.74, and 1.13 points respectively on Chinese-to-English summarization. Moreover, all pre-training objectives have various degrees of contribution to the results, and the monolingual unsupervised objectives (MLM and DAE) are relatively the most important. This verifies the effectiveness of leveraging unsupervised data in the pre-training.

\textbf{Low-resource scenario.} We sample subsets of size 1K and 10K from the training data of cross-lingual summarization and finetune our pre-trained model on those subsets. Figure~\ref{fig:low-resource} shows the the performance of the pre-trained model and the model trained from scratch on the same subsets. As shown, the gain from pre-training is larger when the size of training data is relatively small. This proves the effectiveness of our approach to deal with low-resource language in cross-lingual summarization.

\section{Conclusion}
We present a mix-lingual pre-training model for cross-lingual summarization. We optimize a shared encoder-decoder architecture for multi-lingual and multi-task objectives. Experiments on a benchmark dataset show that our model outperforms pipeline-based and other end-to-end baselines. Through an ablation study, we show that all pretraining objectives contribute to the model's performance.
 
\bibliography{main}
\bibliographystyle{acl_natbib}




\end{document}